\documentclass[runningheads]{llncs}

\usepackage{eccv}
\usepackage{eccvabbrv}

\usepackage{graphicx}
\usepackage{booktabs}
\usepackage{wrapfig}

\usepackage{multirow}
\usepackage{makecell}
\usepackage{graphicx}
\usepackage{array}
\usepackage[accsupp]{axessibility}
\usepackage{microtype}
\usepackage{tcolorbox}
\usepackage{longtable}
\usepackage{float}
\usepackage{listings}
\usepackage{verbatim}
\usepackage{hyperref}
\usepackage{orcidlink}

\newcommand{\methodname}{I-Design\xspace}

\newcommand{\spm}[1]{\tiny{$\,\pm$#1}}

\begin{document}

\title{I-Design: Personalized LLM Interior Designer}

\titlerunning{I-Design: Personalized LLM Interior Designer Agents}

\author{Ata Çelen\inst{1} \and
Guo Han\inst{1} \and
Konrad Schindler \inst{1} \and
Luc Van Gool \inst{1} \and
Iro Armeni \inst{2}$^*$ \and
\\
Anton Obukhov \inst{1}$^*$ \and
Xi Wang \inst{1}$^*$
}
\authorrunning{Çelen A., \etal}

\institute{
ETH Zürich \and
Stanford University \\
$^*$ Equal supervision
}

\maketitle

\begin{center}
    \url{https://atcelen.github.io/I-Design/}
\end{center}

\begin{abstract}
Interior design allows us to be who we are and live how we want -- each design is as unique as our distinct personality.
However, it is not trivial for non-professionals to express and materialize this since it requires aligning functional and visual expectations with the constraints of physical space; this renders interior design a luxury.
To make it more accessible, we present \textbf{\methodname}, a personalized interior designer that allows users to generate and visualize their design goals through natural language communication. 
\methodname starts with a team of large language model agents that engage in dialogues and logical reasoning with one another, transforming textual user input into feasible scene graph designs with relative object relationships.  
Subsequently, an effective placement algorithm determines optimal locations for each object within the scene. 
The final design is then constructed in 3D by retrieving and integrating assets from an existing object database. 
Additionally, we propose a new evaluation protocol that utilizes a vision-language model and complements the design pipeline. 
Extensive quantitative and qualitative experiments show that \methodname outperforms existing methods in delivering high-quality 3D design solutions and aligning with abstract concepts that match user input, showcasing its advantages across detailed 3D arrangement and conceptual fidelity.
\end{abstract}

\section{Introduction}
\label{sec:intro}

Our lives are intimately connected to the spaces we inhabit. Whether renting or buying, these spaces become the backdrop for our memories, hobbies, and time with loved ones. However, finding or creating the perfect space to match our lifestyle, aspirations, and needs is not always straightforward, and professional assistance can be considered a luxury. Even when seeking help from experts, the gap between what inhabitants truly desire and their ability to convey it in the language of professionals often leads to unsatisfactory results. This discrepancy can leave people with ill-fitting living spaces, impacting their physical and mental well-being. Designing interior spaces better suited for individual needs should be accessible to everyone. 

\begin{figure}[t!]
  \centering
  \includegraphics[width=1.0\textwidth]{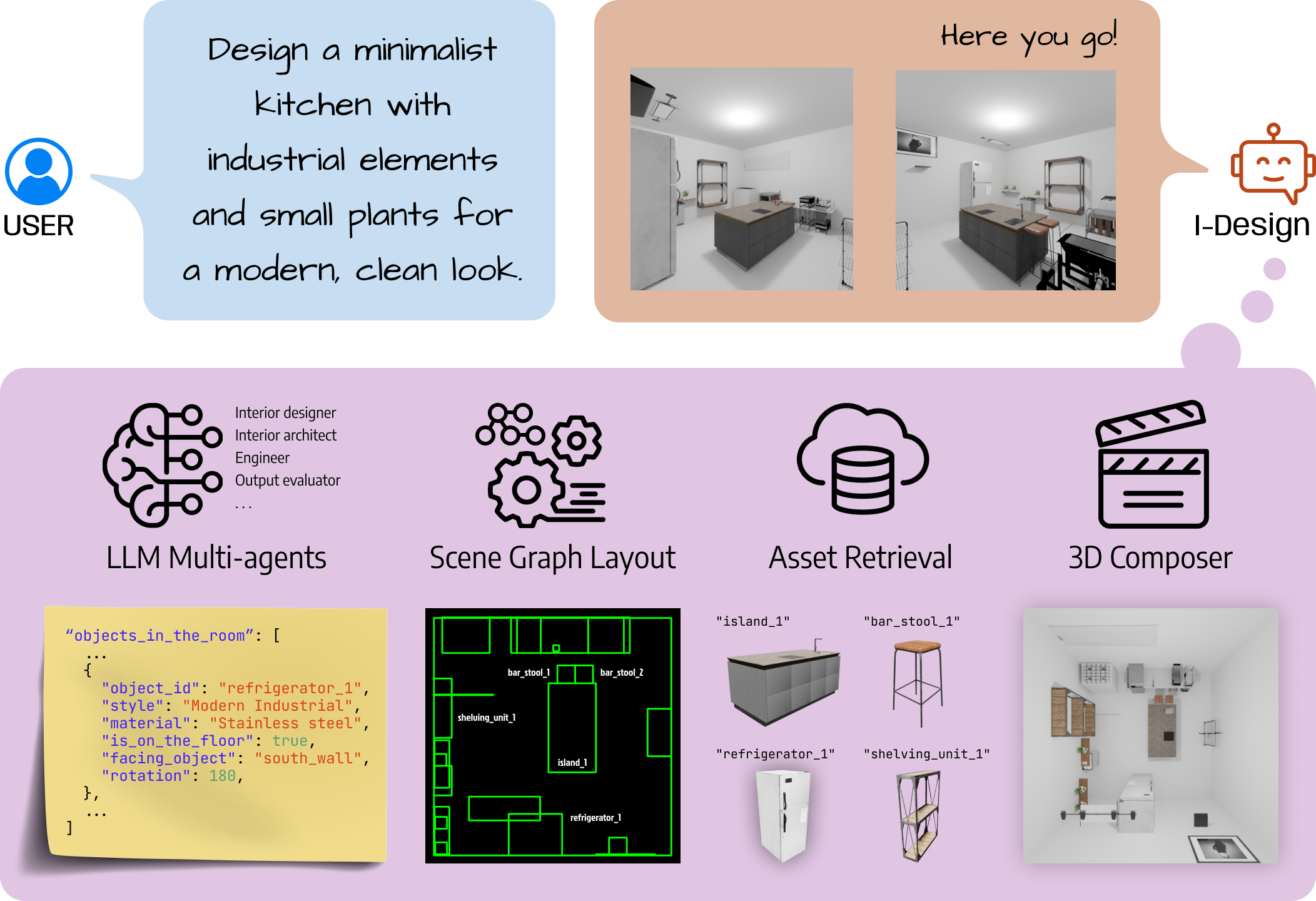}
  \caption{
    \textbf{Overview of \methodname.}
    Starting from the user specification of design preferences in plain text, we query LLM agents to come up with room items, their properties, and their relative relationships in the form of a scene graph. We solve for absolute object placement in the scene graph using the proposed backtracking algorithm (Scene Graph Layout), retrieve 3D assets according to the functional and stylistic specifications, and compose the final result in 3D.
  }
  \label{fig:overview}
\end{figure}

Toward this, we tackle the challenging task of 3D Indoor Scene Synthesis (\textbf{3DISS}). Given a user's unstructured textual description of preferences, we aim to deliver 3D design solutions that align with the latter. 

Specialized generative models~\cite{hollein2023text2room, chung2023luciddreamer, Tang2023mvdiffusion} and data-driven approaches~\cite{wei2023lego, tang2023diffuscene, zhai2023commonscenes} have demonstrated a remarkable ability to produce diverse and realistic interior layouts. 
Yet, their performance is governed by the closed-set and limited datasets used to train them, which inevitably is a biased and incomplete sample of the world. In addition, such methods that accept textual input from the user can only utilize structured text with predefined grammar and rules.
Consequently, it is challenging to guide these models toward producing practical designs for real-world, unseen interiors that align with user preferences. 

The 3DISS task requires reasoning abilities beyond any specialized data-driven model. This includes understanding design principles and concepts like object selection, styles, and spatial arrangement. However, a potential solution must succeed at several fundamental steps to become successful in 3DISS. Firstly, the underlying generator must interpret the abstract input proficiently, \ie identifying the objects to incorporate into the scene to meet user preferences. Simultaneously, it should know about the typical items associated with specific room types, incorporating high-frequency and common-sense objects even if the user does not explicitly mention them. In addition, it requires an awareness of plausible spatial relationships among objects originating from an unrestricted vocabulary. Addressing these challenges requires a system with a comprehensive understanding of diverse human interior design preferences and an extensive database of 3D objects, encompassing their functionality, dimensions, and stylistic attributes. 

Given recent technological advancements, the most viable option to overcome these obstacles is large language models (LLMs) \cite{yang2023harnessing, achiam2023gpt}.
LLMs are trained on internet-scale data and encode a ``world model'' that can be probed and interacted with using natural language.
Still, they cannot use the language as humans do to solve complex tasks without resorting to specialized techniques, such as Chain-of-Thought~\cite{wei2022chain}. 
Technical aspects (\eg limited context window length, number of parameters) and artifacts (\eg hallucinations) additionally hamper practical applications. 
Attempts to produce structured output with LLMs without incorporating hard constraints, communication interfaces, and cross-checks are exciting at first glance but leave much room for improvement. 
LayoutGPT~\cite{feng2023layoutgpt}, a recent work that utilizes LLMs for 3DISS,  directly predicts the absolute positions of objects in the scene. Although this technique may be effective when dealing with small-scale arrangements of a few objects, it proves inadequate in generating realistic scenes containing dozens of objects interlinked in intricate ways. Moreover, the single-step scene generation in LayoutGPT cannot provide interpretability regarding the resulting objects and their arrangement.

Recent research \cite{yang2023auto, qian2023communicative} has also demonstrated that when multiple LLM agents with diverse responsibilities communicate with each other, they can collectively tackle complex tasks with which a single LLM instance may struggle. Hence, we utilize LLM agents to address the challenges posed by 3DISS, leading to the creation of scenes that are spatially more plausible and diverse. Furthermore, drawing inspiration from other works \cite{armeni20193d, chang2015text, zhai2023commonscenes, wang2019planit}, we employ scene graphs as scene representations since they offer a high-level abstraction by focusing on the objects and their relationships, which can be creatively developed with LLMs, refined through rule-based feedback, and visualized. Additionally, employing the scene graph representation as the interface between LLM agents and algorithms enables interpretable object arrangements.

To this end, we present \textbf{\methodname}, a personalized interior designer for 3D Indoor Scene Synthesis. Starting from the user specification of the design preferences in plain text, \methodname queries LLM agents to come up with room items, their properties, and relative relationships in the form of a scene graph. It solves for absolute object placement in the scene graph using the proposed backtracking algorithm, retrieves 3D assets according to the functional and stylistic specifications, and composes the final result in 3D. 
To evaluate the proposed design pipeline, and following in the footsteps of~\cite{wu2024gpt}, we propose a novel evaluation protocol based on a vision-language model (VLM). 
Extensive quantitative and qualitative experiments show that \methodname outperforms existing methods in delivering high-quality 3D design solutions that align with abstract concepts in the user input. 

Our contributions are summarized as follows:
\begin{itemize}
    \item A novel method that takes an unstructured, grammar-free natural language user input and provides 3D design solutions that align with user preferences.
    \item A new approach to the 3DISS task through the reasoning and conversation of multiple LLM agents.  
    \item A procedural scene graph layout transformation, converting scene graphs with relative node relationships into final absolute 3D representations. 
    \item An interpretable pipeline, providing flexibility and enabling iterative design without redoing the entire process. 
    \item A VLM-based evaluation for 3D scenes.
\end{itemize}

\section{Related Work}

\paragraph{\textbf{3DISS via LLMs}:}
Advancements in LLMs have already impacted 3D scene synthesis, even though the connections between the two have not been fully explored.
Initial scene synthesis methods integrated LLMs to encode user textual input into vector representations that were subsequently used in the object placement process \cite{tang2023diffuscene, wang2021sceneformer, lin2024instructscene, liu2023cliplayout}. 
Feng \etal \cite{feng2023layoutgpt} expanded the scope with LayoutGPT by exclusively employing GPT models to generate 3D indoor scene representations. It functions like a retrieval system, employing a strategy based on absolute coordinates to position objects in the scene. Wen \etal \cite{wen2023anyhome} further advance this line of research by developing a dataset-free, open-vocabulary approach for generating 3D home layouts. 
Such endeavors often encounter challenges stemming from the limitations of existing GPT models regarding geometric reasoning, resulting in scenes with objects that overlap or are placed outside the scene boundaries. 

The concept of specialized multi-agents, introduced through projects like AutoGPT \cite{yang2023auto} and ChatDev \cite{qian2023communicative}, has already found diverse applications in data analysis \cite{jin2023genegpt},  interactive reasoning \cite{lin2023swiftsage, wang2023jarvis}, software development \cite{qian2023communicative}, and planning \cite{song2023llm, chen2023autoagents, shen2023hugginggpt}. Our work addresses the above 3DISS challenges by reasoning with LLMs in a multi-agent setting.

\paragraph{\textbf{3DISS via Generative Models:}}
One formulation of 3DISS involves generating multi-view consistent image sets or panoramas according to user-defined specifications and converting these 2D representations into 3D scenes. Approaches like Text2Room \cite{hollein2023text2room} and LucidDreamer \cite{chung2023luciddreamer} use a virtual camera to navigate the space and iteratively generate images through image inpainting, monocular depth estimation, 3D lifting, and stitching. Alternatively, MVDiffusion \cite{Tang2023mvdiffusion} utilizes correspondence-aware attention modules to generate multi-view consistent panoramas or image sets in a single pass. Notably, efforts \cite{tseng2023consistent, shen2023make, bautista2022gaudi,fridman2024scenescape, yu2023wonderjourney,cai2022diffdreamer,wang2023breathing} focusing on consistent novel view synthesis are also adaptable for 3DISS.

Despite leveraging the capabilities of 2D generative models \cite{ho2020denoising, dhariwal2021diffusion, zhang2023adding, rombach2022high}, current methodologies struggle to integrate 3D geometric constraints that are essential for practical interior design applications, such as floor plans and walls. 
Ctrl-Room \cite{fang2023ctrl} attempts to address this challenge, but artifacts persist.
Besides, monocular image-based depth estimation~\cite{ke2023repurposing} and 3D lifting introduce uncertainty into the final 3D mesh output \cite{hollein2023text2room}. Furthermore, the loose semantic coupling across views poses challenges in indoor scene synthesis, leading to unrealistic scenarios like multiple beds in one bedroom \cite{fang2023ctrl, Tang2023mvdiffusion}. The LLM-based approach we employ can also generalize to different settings, such as room type or object descriptions, while simultaneously ensuring the satisfaction of geometric constraints.

\paragraph{\textbf{3DISS through Prior Learning}:}
Another data-driven approach to tackling 3DISS divides the process into two distinct steps: 3D asset selection and layout synthesis, \ie determining the furniture sets for a room and their placements.

For 3D asset selection, two primary approaches emerge: (1) dataset-based retrieval and (2) synthesis via generative models. With the availability of extensive and high-quality asset datasets \cite{deitke2023objaverse, fu20213d, chang2015shapenet, wu20153d}, one can retrieve suitable models, for example, using semantic embedding \cite{liu2024openshape}. Alternatively, generative models can generate appropriate 3D assets based on text or image inputs \cite{poole2022dreamfusion, long2023wonder3d, hong2023lrm, liu2024one}. Given the richness of existing large-scale 3D asset repositories within our context and the direct real-world applicability of retrieving objects from product databases, we use the retrieval-based method.

For furniture arrangement, classical approaches apply specific rules to guide the placement of furniture \cite{weiss2018fast, deitke2022️, xu2002constraint}, construct grammars for procedural modeling \cite{qi2018human, purkait2020sg, kar2019meta}, or use human interaction for editing \cite{merrell2011interactive, huang2023aladdin}. However, with the advent of large-scale 3D indoor datasets \cite{fu20213d, song2017semantic, Structured3D, InteriorNet18}, recent works have shifted towards learning from expert-designed layouts by employing generative models.
Pioneering efforts like ATISS \cite{paschalidou2021atiss} and SceneFormer \cite{wang2021sceneformer} employ transformer-based models to synthesize indoor environments in an autoregressive manner autonomously, selecting and placing objects sequentially. LEGO-NET \cite{wei2023lego} refines initial coarse room layouts by learning human criteria for regularity through a transformer-based diffusion-like pipeline. Recent works like DiffuScene \cite{tang2023diffuscene} and Commonscenes \cite{zhai2024commonscenes} utilize diffusion models to generate interior scenes, representing the scene as a scene graph. To untie layout synthesis from the dataset constraints, we choose to employ LLM-based methods. The common-sense knowledge learned by LLMs can assist in creating a wide range of reasonable designs.

\section{Method}

Our method for proposing design solutions from user input in plain text (\methodname) is summarized in Fig.~\ref{fig:overview}. 
Initially, \methodname examines unstructured textual user input and transforms it into a viable design proposal, represented as a scene graph, through querying LLM agents (Sec.~\ref{sec:llm-agent}). 
We then introduce the scene graph layout module for producing object placement proposals represented by a 2D room floor plan (Sec.~\ref{sec:scene_graph_gen}). 
Finally, we retrieve 3D assets from existing databases (Sec.~\ref{sec:retrieval}) and assemble the final design within a 3D environment, producing functional and stylistic design solutions that reflect user input (Sec.~\ref{sec:composer}). 

\subsection{Problem Formulation}

Given a non-structured textual user input $T_{user}$, the dimensions of a room \begin{math} (l_{room}, w_{room}, h_{room}) \in \mathbb{R}^3  \end{math}, and the number of objects to include in the scene $n \in \mathbb{R}$, the objective is to create a 3D scene that aligns with the free-form user's requests in the textual input, while ensuring functionality, coherent design, and 3D consistency.

We use a scene graph \(\mathcal{G} = (\mathcal{O}, \mathcal{E})\) to represent the spatial relationship between objects, where nodes $\mathcal{O} = [\mathbf{o}_1, ..., \mathbf{o}_n]$ represent object instances in the scene. 
An additional node type \(\textbf{o}^{(r)}\) is included for room layout elements such as walls, ceiling, or the floor. 
Each object $\mathbf{o}_i$ is associated with a set of properties $\mathbf{o}_i = \{\alpha_i, m_i, s_i, r_i, p_i, cs_i\}$, where \(\alpha_i\) denotes the object's name, \(m_i\) describes its material and architectural style. 
The associated geometric properties are \(s_i \in \mathbb{R}^3\) the dimensions of its bounding box, \(r_i \in \mathbb{R}\) the rotation angle along the z-axis pointing upwards, and \(p_i \in \mathbb{R}^3\) the position of the object in the scene.
Finally, \( cs_i = (x_{neg}, x_{pos}, y_{neg}, y_{pos})  \in \mathbb{R}^4 \) is the size of the subgraph's bounding box, indicating the overall space that an object, along with its children, would occupy in each direction.
Edges in the graph are represented by \(\mathcal{E}\) where \( e_{ij} = (\text{adj}_{ij}, \text{prep}_{ij}) \) is the directed edge from $\mathbf{o}_i$ to $\mathbf{o}_j$ that comprises details regarding the adjacency of the two objects and the prepositional connection between them.
$\mathbf{o}_i$ is considered as the parent node to the child node $\mathbf{o}_j$. 
These edges describe the spatial locations between objects (e.g. \textit{left}/\textit{right}, \textit{in front}/\textit{behind}, \textit{on}, and \textit{above}/\textit{under}) as well as the connections to the room layout (e.g. \textit{on} and \textit{in the corner}).

\subsection{LLM Multi-agents}
\label{sec:llm-agent}

\begin{figure}[t!]
  \centering
  \includegraphics[width=1.0\textwidth]{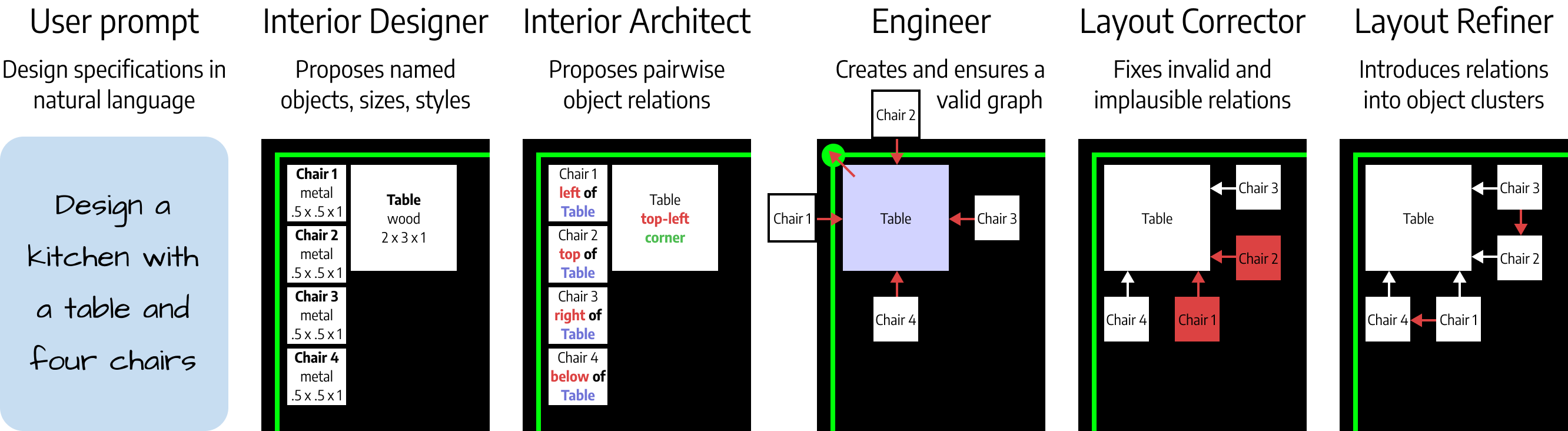}
  \caption{
    \textbf{Scene Graph Generation Pipeline with LLM Agents.}
    Each agent receives the user input and a JSON representation of the previous stages and transforms the output according to the task prompt.
    Due to the specialized nature of each agent, the generated scene graph reflects user specifications and feasibility constraints, such as topological correctness and semantic plausibility.
    Red indicates changes introduced by each agent.
    Complete prompts of every agent are specified in the Appendix.
  }
  \label{fig:agents}
\end{figure}

Generating a 3D indoor scene based on unstructured user input is a challenging task that demands detailed planning. 
Two essential steps are shared among common methods: selecting the objects to populate the scene based on user input and arranging them in a meaningful and coherent configuration. 
We employ a multi-agent approach to tackle this complexity effectively, allocating the tasks involved in these steps across multiple LLM agents. 
As shown in Fig.~\ref{fig:agents}, interpreting and transforming user input into a functional, personalized scene graph involves five distinct agents: Interior Designer, Interior Architect, Engineer, Layout Corrector, and Layout Refiner. 
The Interior Designer and Architect propose a relative scene graph from the user input. We assume that the agents have appropriately considered contextually bonded objects while constructing the relative scene graph, justifying spatial separation and disconnection within the graph. 
However, the relative scene graph lacks guaranteed spatial consistency, allowing for connections between objects that may defy real 3D scene constraints.
Therefore, our pipeline uses a Layout Corrector agent to rectify invalid and implausible connections, while the scene graph contains only relative relations.
Finally, to increase the pipeline's robustness when dealing with multiple objects assigned to the same parent within the scene graph, the Layout Refiner agent proposes relations between such objects, simplifying the subsequent layout stage. Next, we examine each of the agents in more detail.

\textbf{Interior Designer:} 
The agent receives inputs comprising the free-form user input $T_{user}$, the room dimensions $(l_{room}, w_{room}, h_{room})$, and the desired number of objects $n$, and proposes a selection of objects $\{\mathbf{o}_i\}$ tailored to the user's preferences, while also ensuring that their functionality matches the room type. 
Specifically, for each object, the interior designer suggests the following properties
$\mathbf{o}_i = \{\alpha_i, m_i, s_i, r_i, p_i\}$ including its name $\alpha_i$, material $m_i$, 3D size $s_i$, orientation $r_i$, and position $p_i$. 
Note that the agent may propose several identical instances. 
For example, there might be four instances of the ``chair'' type positioned around a table (see Fig.~\ref{fig:agents}).

\textbf{Interior Architect:} 
Its primary responsibility is to establish object-to-object connections, as well as object-to-room layout relations. 
In other words, its role is to establish the edge connections $e_{ij}$ between the object nodes. 
The agent is not constrained on the number of edges each object can have, allowing for diverse configurations where an object may have edges to room layout elements but none to other objects, and vice versa. 
Additionally, the Interior Architect creates information on the rotation $r_i$ of each object around the vertical axis. 

\textbf{Engineer:}
This agent transforms the relative scene graph into a JSON object structured according to a specified schema. Each entry in the JSON file encompasses the details for $\mathbf{o}_i = \{\alpha_i, m_i, s_i, r_i, p_i\}$ for the corresponding object. Moreover, the agent employs a JSON schema validator that assesses the validity of the generated file based on the provided schema. In cases of non-compliance, a modification from the Engineer is required 
until the output aligns with the specified schema, ensuring a valid JSON representation.

\textbf{Layout Corrector:} 
The responsibility of this agent is to fix invalid connections in the graph, which involves removing spatial implausible edges and eliminating ambiguities between nodes. 
We preemptively examine three types of spatial implausibilities: (1) Checking for objects positioned beyond room boundaries (as exemplified in Fig.~\ref{fig:agents}), (2) Assessing spatially impossible object-to-object connections, and (3) Ensuring size compatibility between parent and child objects.
The agent determines the relocation of objects experiencing spatial conflicts, either suggesting alternative edges for these objects or removing them entirely from the scene.

\begin{enumerate}
\item To check room boundaries, we examine children nodes of objects allocated along walls or in room corners.
For example, if object \textbf{A} is positioned \textit{behind} object \textbf{B}, which is placed alongside a wall, this would result in object \textbf{A} being positioned out of bounds.
 \item Assessing the plausibility of object-to-object connections involves verifying whether adjacency relations between objects could be violated via conflicting edges in the scene graph. This requires checking if any object has been positioned between two adjacent objects in the scene graph.
\item We operate under the assumption that, in a valid scene graph, the bounding box of the parent object is sufficiently spacious to accommodate its children.
For example, if a \textit{table} has two \textit{chairs} positioned on its \textit{left} side, it should be wide enough to accommodate them; likewise, a \textit{wall} with multiple objects attached \textit{on} it should have sufficient size to accommodate all of its children objects. 
The last check ensures this compatibility. 
\end{enumerate}

\textbf{Layout Refiner:} 
The agent aims to eliminate ambiguities between children nodes that share the same edge from the parent. A notable example of this phenomenon is observed with ornaments placed on a desk. As a desk typically accommodates several objects on its surface, a scene graph designating the \textit{desk} as the parent with the preposition \textit{on} as the edge fails to convey the relative orientation of the objects on the desk to each other. To eliminate this ambiguity, we establish edges between children nodes, ensuring a distinct ordering among them. 
The agent also verifies that the scene graph maintains its acyclic property. If the graph contains cycles, we must remove edges contributing to these cycles to preserve a distinct hierarchy within the scene graph.

\subsection{Scene Graph Layout}
\label{sec:scene_graph_gen}

\subsubsection{Computing Cluster Dimensions} 
\label{cluster_sizes}

In the final postprocessing phase, we compute cluster dimensions for each object to facilitate the positioning in the subsequent backtracking algorithm, thereby strategically constraining the search space. Leveraging that \(\mathcal{G}\) is a Directed Acyclic Graph, we perform a topological sort on the nodes to establish a hierarchical ordering. This approach enables us to determine the clearance -- the distance an object must spare in each direction to ensure the successful placement of its children within the scene.

Taking such precaution becomes crucial in scenes where the search space for each object is extensive, yet the solution space is relatively constrained. Without this precaution, the randomized nature of the backtracking algorithm may lead to extended processing times, particularly when dealing with many object relationships in the scene.

\subsubsection{The Backtracking Algorithm} 
\label{backtracking}

The backtracking algorithm serves to convert the relative representation of the scene graph \(\mathcal{G}\) into object positions \(\mathcal{P}\). The room is initially populated with the root nodes of the scene graph \(\mathcal{G}\), representing the fundamental elements of the room layout, such as walls and ceiling. The positions of these root nodes remain fixed while other objects are arranged around them. 
Conceptually, the algorithm represents the plausible position of each object as a bounding box \(\mathcal{B}_i\) and sampling a point \(p_i\) for each \(\mathbf{o}_i\) from \(\mathcal{B}_i\).

The objects are placed after being topologically sorted, prioritizing nodes higher in the scene graph hierarchy for placement first. To avoid placement issues later in the algorithm, each object \(\mathbf{o}_i\) is positioned alongside its children according to \(\mathcal{G}_i\). 
The plausible positions bounding box for the objects is defined through predetermined placement functions \( f(\mathbf{o}_i, \mathbf{o}_j)\) \( \forall \mathbf{o}_j, \mathbf{o}_i \in \mathcal{G}_j \). 

\begin{figure}[t!]
  \centering
  \includegraphics[width=1.0\textwidth]{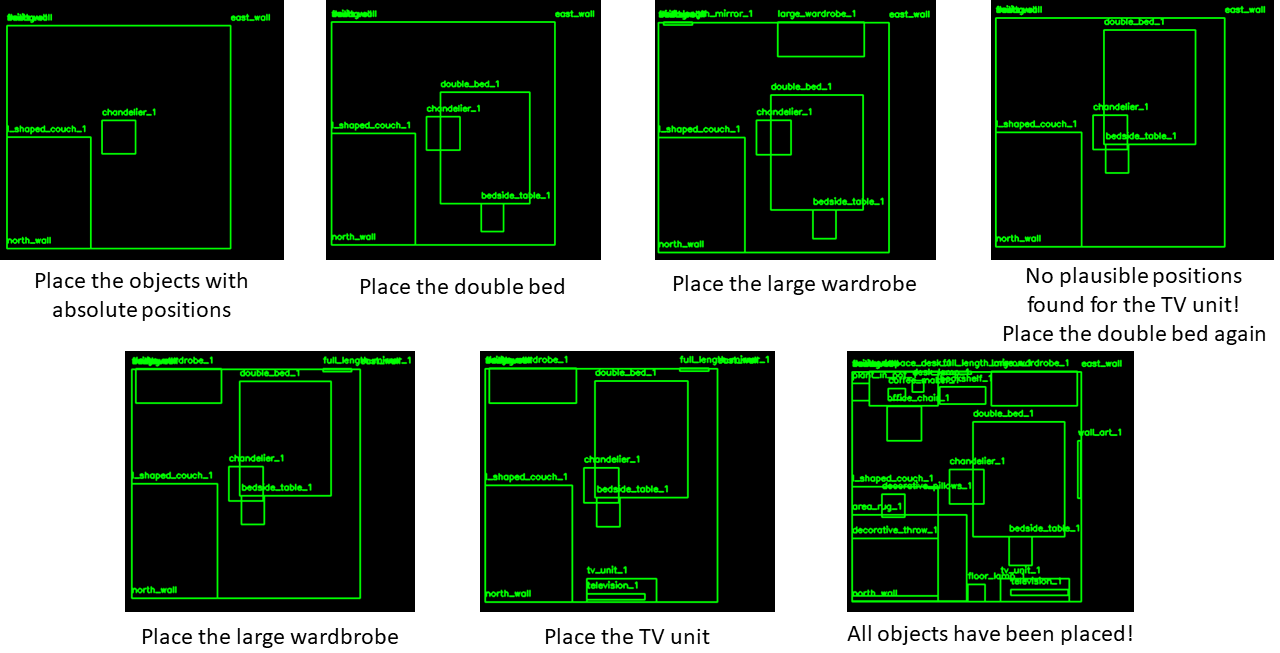}
  \caption{
    \textbf{Backtracking-based Scene Graph Layout.}
    Starting from the scene graph specifying interior items as nodes connected with relative relationships, the backtracking algorithm solves for their absolute placement in the room. The algorithm reverses dead-end configurations and excels at object placement while respecting spatial constraints.
  }
  \label{fig:backtracking}
\end{figure}

The nodes are organized into depth groups, where a node's depth \(d = |\mathbf{o}_i \rightarrow \textbf{o}^{(r)}|\) represents its distance to the closest \(\textbf{o}^{(r)}\). If an object \(\textbf{o}_i\) with depth \(d\) cannot be placed into the scene due to either an empty bounding box \(\mathcal{B}_i = \varnothing\) or if \(p_i\) consistently results in collisions with other objects, we remove \(p_j\) for all objects with depth \(\geq d\). Subsequently, we re-sample positions for objects with depth \(d-1\). Once all objects with depth \(d\) are positioned successfully, we increment the depth counter of the algorithm. An example process of scene graph processing is shown in Fig.~\ref{fig:backtracking}.

\subsection{3D Asset Retrieval} 
\label{sec:retrieval}

We can generate textual descriptions for each object by utilizing the object name, style, and material information. 
These textual descriptions are then transformed into text embeddings using the CLIP \cite{radford2021learning} text encoder. 
The alignment between OpenShape \cite{liu2024openshape} encodings and CLIP embeddings enables us to measure the distance between our text embeddings and the learned object representations from the database of choice. 
This facilitates the retrieval of a 3D asset that is closest to our textual description of each object. 
After the retrieval, the assets can be adjusted to fit the bounding box provided to these objects within the scene graph.

\subsection{3D Composer} 
\label{sec:composer}

Having produced the final scene graph and retrieved the 3D assets, we employ an off-the-shelf 3D renderer to visualize the room interior in 3D.
The output of \methodname pipeline is a collection of entities, including (1) the scene graph, (2) the floor plan, (3) preconfigured rendered views, and (4) interpretability artifacts, such as the input user prompt and the communication log between the agents.

This output bundle serves multiple purposes. 
Primarily, it allows for the inspection of every pipeline stage and, if necessary, enables the replay of select stages to introduce variations. 
For example, the user may want to swap out a piece of furniture or render the scene from a novel viewpoint. 
Secondarily, the bundle enables diverse quantitative studies to investigate the alignment of user preference with the produced results. 
We elaborate on this aspect in Sec.~\ref{sec:quantitative}.

\section{Experiments}
\subsection{Implementation Details}

We use Microsoft's AutoGen~\cite{wu2023autogen} framework to enable multi-agent conversation, with each agent equipped with GPT-4 model~\cite{achiam2023gpt}. The \texttt{temperature} is configured at 0.7, while \texttt{top\_p} is set to 1.0. For object retrieval, we rely on OpenShape~\cite{liu2024openshape}, utilizing text embeddings to retrieve objects from Objaverse~\cite{deitke2023objaverse}. The scene is then visualized and rendered using Blender~\cite{blender}.

\subsection{Quantitative Evaluation}
\label{sec:quantitative}

We compare \methodname with baseline approaches to quantitatively assess the design quality of the interior rooms generated by our proposed pipeline.

\subsubsection{Metrics}
\label{sec:metrics}

We conduct quantitative evaluations using the following metrics:

\noindent
\textit{Average Number of Proposed Objects (NObj): } 
The diversity of the generated interior rooms can be partially assessed by considering the number of furniture pieces proposed and placed within the scene. Here, we compute the average number of proposed objects separately for bedrooms and living rooms. 

\noindent
\textit{Out-of-Boundary Rates (OOB): } 
The feasibility of a generated layout can be quantified by examining the frequency of objects extending beyond room boundaries, as outlined in~\cite{feng2023layoutgpt}. In this context, if any bounding boxes are outside the designated room boundaries within a scene, that scene is deemed invalid. OOB values are determined by calculating the ratio of invalid scenes to the total number of generated scenes.

\noindent
\textit{Bounding Box Loss (BBL):} 
The viability of a generated layout can also be measured by evaluating the degree of overlap between proposed furniture bounding boxes. This is calculated as the average volume of bounding box intersections across generated scenes.

\noindent
\textit{GPT-4V ratings:} 
As highlighted in~\cite{wu2024gpt}, GPT-4V~\cite{achiam2023gpt} has been shown to serve as a human-aligned evaluator for 3D content. Inspired by this concept, we employ GPT-4V as an evaluator to assess the visual quality of synthesized rooms based on their renderings. The rating criteria encompass aspects of functionality and activity-based alignment (Func.), layout and furniture (Layout.), color scheme and material choice (Scheme.), as well as overall aesthetic and atmosphere (Atmos.). The integer ratings range from 0, the minimum, to 10, the maximum. The rating process is conducted unary, where each scene is evaluated individually. Two renderings from different viewpoints are horizontally concatenated and passed along with the user input to the evaluator for rating to quantify their alignment with the generated results.
Each evaluation is performed three times in parallel by setting the corresponding parameter \texttt{n} using the GPT-4V API. 
The resulting grades are averaged in every category, and their standard deviations are also reported to account for the grade spread and allow for pairwise grade comparisons.

\subsubsection{Baseline}

We treat LayoutGPT~\cite{feng2023layoutgpt} as a baseline for comparison. LayoutGPT~\cite{feng2023layoutgpt} utilizes LLM models, including GPT-4~\cite{achiam2023gpt}, to generate room layouts in a CSS-like format through few-shot learning with examples sourced from the 3D-FRONT~\cite{fu20213df} dataset. 
Here, for a fair comparison, we have extended the original LayoutGPT pipeline~\cite{feng2023layoutgpt} by enabling it to retrieve and place objects from Objaverse~\cite{deitke2023objaverse}.

\begin{figure}[ht!]
  \centering
  \includegraphics[width=1.0\textwidth]{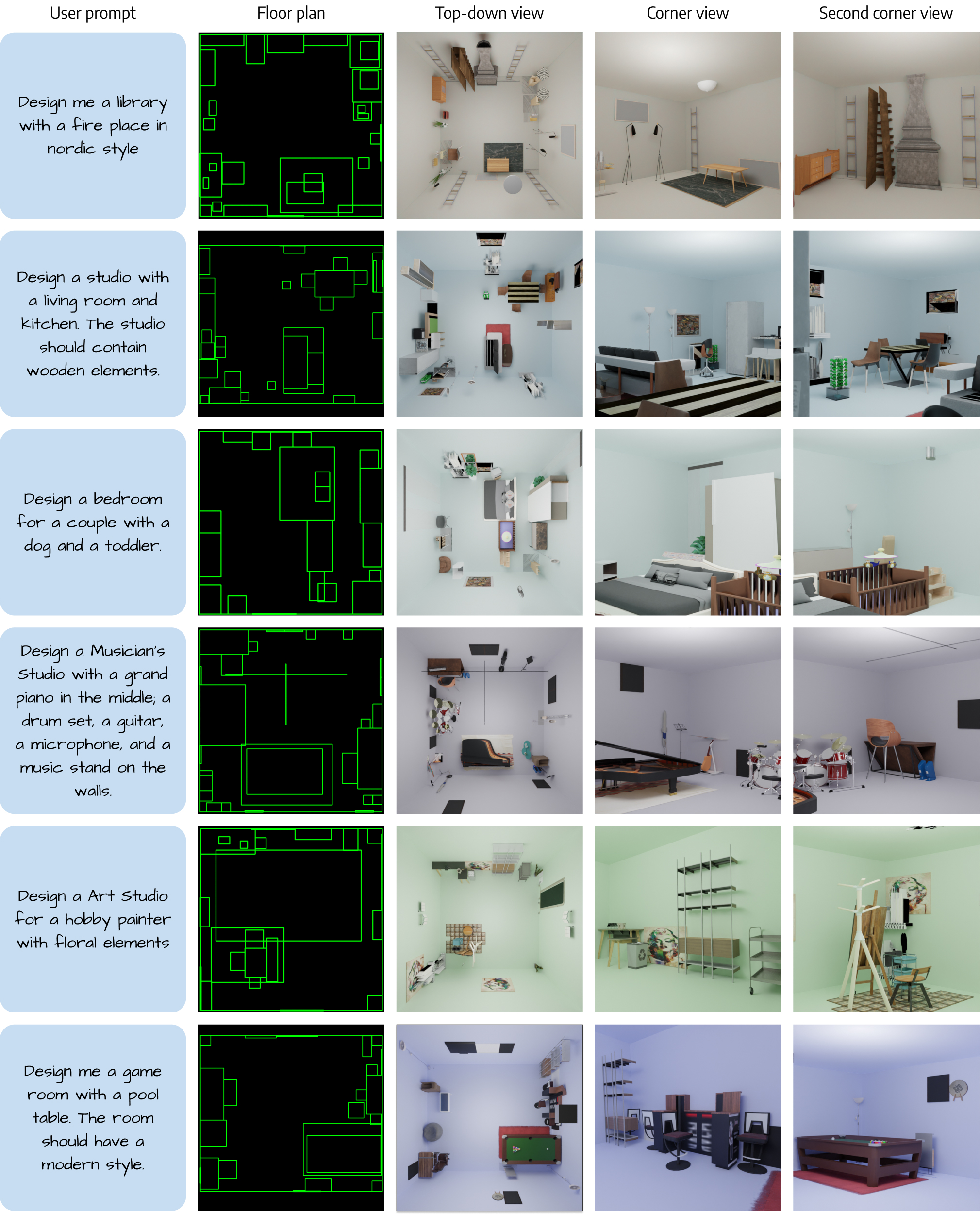}
  \caption{
    \textbf{Gallery of Results Obtained with \methodname.}
    The first column lists user prompts, specifying design preferences, such as functionality, style, and atmosphere. 
    The top-down scene graph layout generated by the LLM Agents
    is shown in the second column.
    The third column shows this layout rendered from the top-down view after the Asset Retrieval stage.
    The last two columns show corner views of the resulting design.
    Evidently, \methodname is capable of factoring in diverse user specifications and producing vibrant, functional designs.
  }
  \label{fig:gallery}
\end{figure}

\subsubsection{Experimental Settings}

As LayoutGPT~\cite{feng2023layoutgpt} only supports ``Bedroom'' and ``Living room'' room types, our comparison is limited to these types and variations of room dimensions. 
We ensure that the prompts given to our pipeline contain only this information for a fair comparison. 
Besides, we let LayoutGPT~\cite{feng2023layoutgpt} retrieve objects from Objaverse~\cite{deitke2023objaverse} and use the same settings for rendering. 
We generate ten living room and ten bedroom scenes with the same dimensions for each method. The quantitative comparison results are provided in \cref{tab:quant_eval}.

\begin{table}[tb]
\caption{
\textbf{Quantitative comparison with LayoutGPT~\cite{feng2023layoutgpt}.}
Both methods generate 10 ``Bedroom'' and 10 ``Living room'' scenes of varying room sizes.
The objective metrics are:
an average number of objects (NObj), out-of-bounds rate (OOB), and bounding box loss (BBL). 
Subjective metrics:
GPT-4V~\cite{achiam2023gpt} is used as a human-aligned evaluator assessing the following scene aspects: functionality and activity-based alignment (Func.), layout and furniture (Layout.), color scheme and material choice (Scheme.), overall aesthetic and atmosphere (Atmos.). 
The evaluator LLM is prompted to evaluate the alignment of the user input with the rendered scene, assign a grade from 0 to 10, and explain the reasoning behind the assignment (Sec.~\ref{sec:metrics}). 
Each scene is evaluated 3 times, resulting in the mean and standard deviation per scene, averaged across scenes.
\methodname produces more realistic scenes, which score better in all metrics with tight confidence.
}
\label{tab:quant_eval}
\begin{center}
\vspace{-2em}
\resizebox{1.0\linewidth}{!}{
\begin{tabular}{
>{\arraybackslash}p{2.5cm} 
>{\arraybackslash}p{2.2cm} 
*{9}{>{\centering\arraybackslash}p{1.2cm}}
}

\toprule

\multirow{2}*{Method} & 
\multirow{2}*{Room Type} &
\multirow{2}*{NObj~$\uparrow$} & 
\multirow{2}*{OOB~$\downarrow$} & 
\multirow{2}*{BBL~$\downarrow$} & 
\multicolumn{5}{c}{\makecell{GPT-4V Criteria~$\uparrow$}} \\

& 
& 
& 
& 
& 
\tiny{Func.~$\uparrow$} & 
\tiny{Layout.~$\uparrow$} & 
\tiny{Scheme.~$\uparrow$} & 
\tiny{Atmos.~$\uparrow$} &
\tiny{Avg.~$\uparrow$} \\

\midrule

\multirow{3}*{LayoutGPT~\cite{feng2023layoutgpt}} & 
Bedroom & 
5.5 & 
51.06 & 
14.09 & 
4.8\spm{0.4} & 
4.8\spm{0.8} & 
4.6\spm{0.8} & 
4.9\spm{0.4} &
4.8 \\

& 
LivingRoom & 
6.9 & 
64.15 & 
1.06 & 
4.8\spm{1.3} & 
4.8\spm{0.8} & 
4.8\spm{1.4} & 
4.6\spm{0.8} &
4.8  \\

& 
Avg. & 
6.2 & 
57.6 & 
7.58 & 
4.8 & 
4.8 & 
4.7 & 
4.8 &
4.8  \\

\midrule

\multirow{3}*{Ours (\methodname)} & 
Bedroom & 
12.7 & 
0.0 & 
0.34 & 
5.2\spm{0.4} & 
5.5\spm{0.2} & 
5.6\spm{0.7} & 
5.5\spm{0.2} &
5.5  \\

& 
LivingRoom & 
23.6 & 
0.0 & 
0.31 & 
5.8\spm{2.6} & 
5.6\spm{2.1} & 
5.9\spm{1.1} & 
5.7\spm{1.3} &
5.8  \\

& 
Avg. & 
\textbf{18.2} & 
\textbf{0.0} & 
\textbf{0.33} & 
\textbf{5.5} & 
\textbf{5.6} & 
\textbf{5.8} & 
\textbf{5.6} &
\textbf{5.7}  \\

\bottomrule

\end{tabular}
        
}
\vspace{-2em}
\end{center}
\end{table}

Besides comparing our approach with the baseline, we aim to demonstrate our ability to appropriately process prompts covering diverse aspects. For this purpose, we generate ten prompts for each potential factor influencing an individual's decisions regarding interior designs, encompassing functionality, layout, color scheme, and overall atmosphere. We compare the rooms synthesized by tailored prompts focusing on specific aspects~\cref{tab:prompt_eval_unary}.
Refer to the Appendix for a complete list of prompts used for both tables.

\subsubsection{Results}

\begin{wraptable}{l}{0.5\textwidth}
\vspace{-3.5em}
\caption{
\textbf{Quantitative evaluation} of \mbox{\methodname} with various prompt types. 
Each prompt type produced 10 instances of a text prompt, each used to generate a single scene. 
These scenes were evaluated with the LLM scheme similarly to~\cref{tab:quant_eval}, and the grades were aggregated across prompt types.
Maximal alignment between visual results and the input user prompt is reached when adding ``Functional'' cues. 
Refer to prompt examples in the Appendix.
}
\label{tab:prompt_eval_unary}
\begin{center}
\vspace{-1em}
\resizebox{1.0\linewidth}{!}{
\begin{tabular}{
*{1}{>{\arraybackslash}p{2cm}}
*{5}{>{\centering\arraybackslash}p{1.2cm}}
}
            
\toprule
            
\multirow{2}*{Prompt type} &
\multicolumn{5}{c}{\makecell{GPT-4V Criteria}} \\

&
\tiny{Func.} & 
\tiny{Layout.} & 
\tiny{Scheme.} & 
\tiny{Atmos.} & 
\tiny{Avg.}  \\

\midrule

Atmospheric &
4.9\spm{0.7} & 
4.9\spm{0.6} & 
4.3\spm{0.6} & 
4.4\spm{0.5} & 
4.9 \\

Scheme &
4.6\spm{0.5} &
4.8\spm{0.6} & 
5.0\spm{0.5} & 
4.8\spm{0.5} & 
5.0 \\

Layout &
6.2\spm{0.6} & 
6.0\spm{0.7} & 
5.2\spm{0.7} & 
5.6\spm{0.6} & 
5.8 \\

Functional &
\textbf{7.0}\spm{0.7} & 
\textbf{6.5}\spm{0.7} & 
\textbf{5.8}\spm{0.8} & 
\textbf{6.2}\spm{0.7} & 
\textbf{6.4} \\

\bottomrule
        
\end{tabular}

}
\vspace{-2.5em}
\end{center}
\end{wraptable}

Compared to LayoutGPT~\cite{feng2023layoutgpt}, as illustrated in \cref{tab:quant_eval}, our approach stands out in proposing larger furniture sets and more physically plausible layouts, exhibiting zero OOB and minimal BBL values. For evaluations with GPT-4V, our method outperforms LayoutGPT~\cite{feng2023layoutgpt} across all metrics, indicating better visual quality of the synthesized rooms and alignment with user preferences.

As outlined in \cref{tab:prompt_eval_unary}, 
the performance of \methodname varies depending on the type of prompt input. 
Our method yields the best results in all four grading aspects when incorporating functional descriptions. 
Functionality prompts provide cues regarding the room type and potential activities occurring within the space. 
The usage of atmosphere and scheme information somewhat diminishes the quality of synthesis. 
We attribute this to our pipeline's lack of consideration for the materials and textures of the walls, ceiling, and floor and the absence of an object re-texturing step. 
This can significantly impact the room's overall ambiance and can be considered a promising future research direction.

\subsection{User Study}
\label{sec:user_study}

\begin{figure}[H]
  \centering
  \begin{subfigure}[b]{\textwidth}
    \centering
    \includegraphics[width=\textwidth,height=\textheight, keepaspectratio]{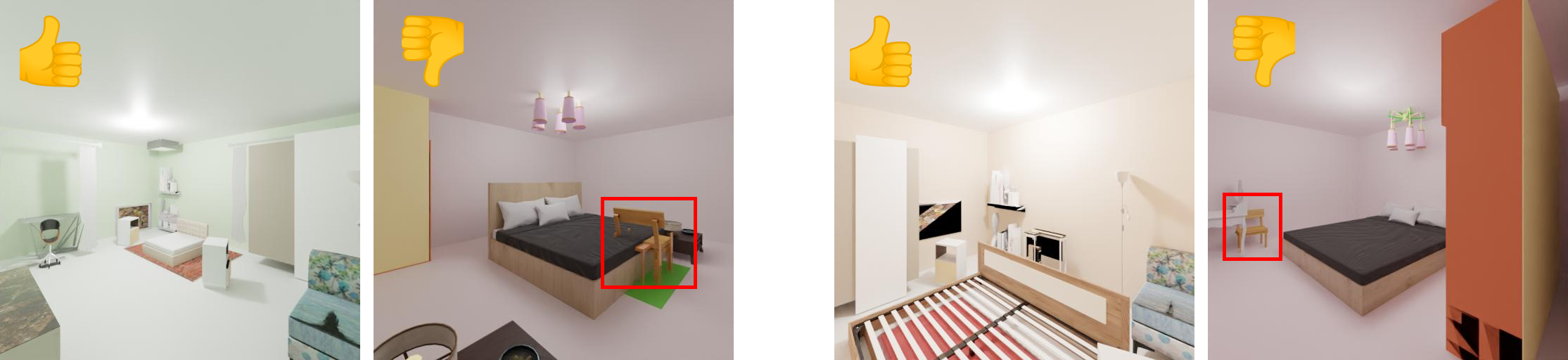}
    \caption{Bedroom Control Scenes}
    \label{fig:userstudybed}
  \end{subfigure}
  \hfill
  \begin{subfigure}[b]{\textwidth}
    \centering
    \includegraphics[width=\textwidth,height=\textheight, keepaspectratio]{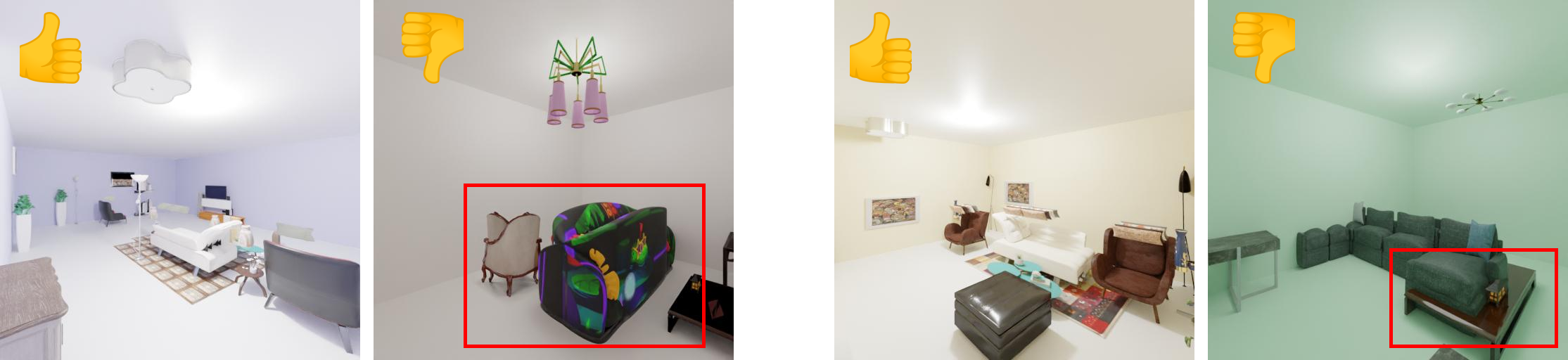}
    \caption{Living Room Control Scenes}
    \label{fig:userstudyliv}
  \end{subfigure}
  \caption{
    \textbf{Control Questions for the Subjective Study}
    Here are two control questions provided for our subjective study focusing on bedrooms in Figure \ref{fig:userstudybed} and living rooms in Figure \ref{fig:userstudyliv}. These questions are designed to prompt participants to assess object collisions and the level of detail present in the scenes. The red boxes show the object collisions in the scenes.
    }
\end{figure}

Tab.~\ref{tab:user_study} contains the results of a user study aimed at reinforcing our
GPT-4V evaluation. 
Out of 20 scenes created by our method and another 20 by LayoutGPT\cite{feng2023layoutgpt}, we randomly sampled 5 scenes from the competitor pool for each sample and thus composed 100 pairs.
Each pair depicts either a bedroom or a living room, each comprising a render from our method and one from LayoutGPT. 
Subjects were instructed~\cite{subjectify} to vote on the most realistic room in each pair. The order of the methods is randomized during the display to prevent subject bias. Each participant was assigned 20 pairs to evaluate, and two control questions with predefined correct answers were included for verification purposes, as shown in Fig.\ref{fig:userstudy}. We collected 1254 answers for bedroom pairs and 660 for living room pairs. The gathered votes were converted into Bradley-Terry preference scores~\cite{bradley1952rank} and probabilities. 
The results affirm our findings with the GPT-4V evaluator, with our method outperforming LayoutGPT for both room types.

\begin{table}[th]
\caption{
\textbf{Subjective Study} of user preferences between LayoutGPT and I-Design. These results confirm the higher realism of our layouts and the reliability of the proposed GPT-4V evaluation scheme.
}
\label{tab:user_study}
\begin{center}
\vspace{-1.5em}
\resizebox{0.8\linewidth}{!}{
\begin{tabular}{
@{}
>{\arraybackslash}p{2.5cm} 
>{\arraybackslash}p{2.5cm} 
>{\centering\arraybackslash}p{3cm} 
>{\centering\arraybackslash}p{2cm}
@{}
}
\toprule

\multirow{1}*{Method} & 
\multirow{1}*{Room Type} & 
\multirow{1}*{Bradley-Terry score~$\uparrow$} & 
\multirow{1}*{Prob.~$\uparrow$} \\

\midrule

\multirow{2} *{LayoutGPT} & 
Bedroom &
0.12 &
0.42 \\

&
LivingRoom & 
0.17 &
0.38 \\

\midrule

\multirow{2} *{Ours} & 
Bedroom &
\textbf{0.40} &
\textbf{0.58} \\

&
LivingRoom & 
\textbf{0.69} &
\textbf{0.62} \\

\bottomrule

\end{tabular}

}
\vspace{-1.5em}
\end{center}
\end{table}

\subsection{Qualitative Study}

The gallery showcasing rooms generated with \methodname provided various prompts is available in \cref{fig:gallery}. Each synthesized room adeptly reflects the specifications from user textual inputs. When users explicitly mention furniture preferences and desired positions, as showcased in the fourth example, the resulting room seamlessly aligns with those requirements. Similarly, when specifications are implicit, as seen in the third example depicting a bedroom tailored for a family with a toddler, the synthesized room's functionality satisfies user preferences, with a crib placed next to the king-sized bed.

\section{Conclusion}

Computing-assisted technologies have been widely used to simplify the interior design process for individuals without expertise. Current methods for generating interior rooms, whether through generative models or prior learning, often suffer from limitations such as inconsistencies between views and restrictions on furniture sets or room types, which are confined to the dataset's provided domain. We aim to eliminate these constraints and provide users with a user-friendly interface that streamlines the construction phase of the interior room. To realize this vision, we presented \methodname.

By leveraging LLM multi-agent architecture, our framework interprets user preferences expressed through text, transforming unstructured text into a structured representation. A furniture set that aligns with the user's requirements is inferred. Subsequently, spatial relationships among furniture are established, with common sense knowledge integrated into the agents' decision-making process. The resulting scene graph provides a visual representation of the proposed design. Crucially, our framework ensures interpretability through a transparent breakdown of each step involved, facilitating an accessible design process. 

Our framework still encounters limitations, notably termination problems where the pipeline fails to find a solution for object placements. This issue arises when handling many objects in a relatively small scene. Spatial conflicts may persist despite efforts in postprocessing to rectify them, or there may not be enough space for furniture placements.
Asset retrieval also poses challenges, as the quality of retrieved assets may vary, and they can even come with invalid default orientations. Additionally, discrepancies in object sizes compared to proposed dimensions may occur, leading to artifacts when resizing them.

A promising future work direction is further exploring object placement with an automated, learning-based approach, moving away from the current backtracking algorithm that relies on trial and error. Another direction would be to adopt a generative approach to replace or complement the object retrieval process, thereby improving the diversity and flexibility of the overall pipeline. 

\bibliographystyle{splncs04}
\bibliography{main}

\clearpage

\begin{center}
\bfseries\Large
I-Design: Personalized LLM Interior Designer:\\
Appendix
\end{center}

\section{Additional Qualitative Results}
\begin{figure}[H]
  \centering
  \includegraphics[width=.80\textwidth,clip,trim={0 60em 0 10em}]{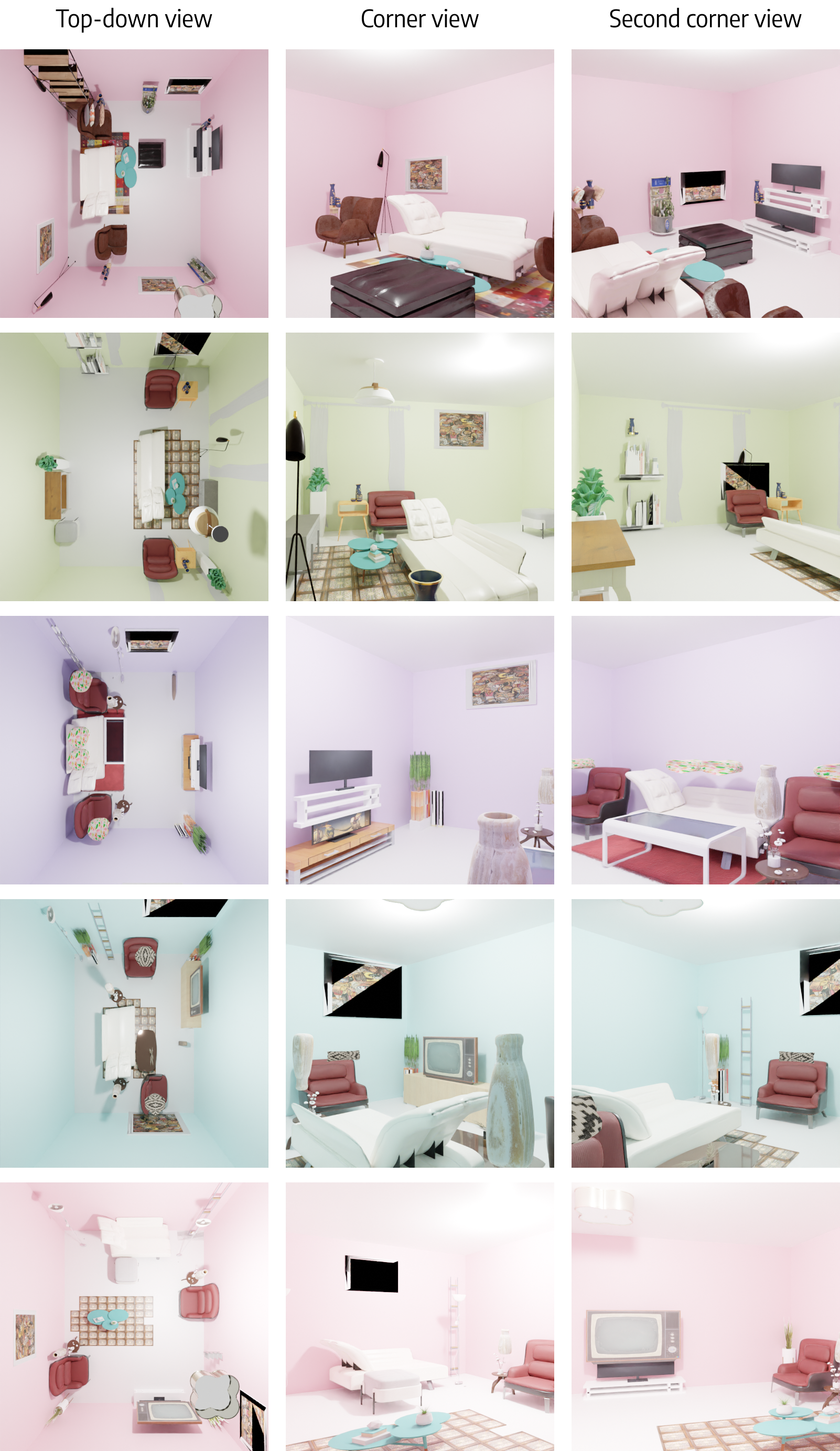}
  \caption{
    \textbf{Living Room Renders with Generic Prompting.}
    We showcase samples from living room renders used for the comparison with LayoutGPT~\cite{feng2023layoutgpt} in Tab.~\ref{tab:quant_eval} of the main paper. 
    The common prompt for all these scenes is: ``\textit{Design me a living room}''.
  }
  \label{fig:supp:gallery}
\end{figure}

\begin{figure}[H]
  \centering
  \includegraphics[width=\textwidth,height=\textheight, keepaspectratio]{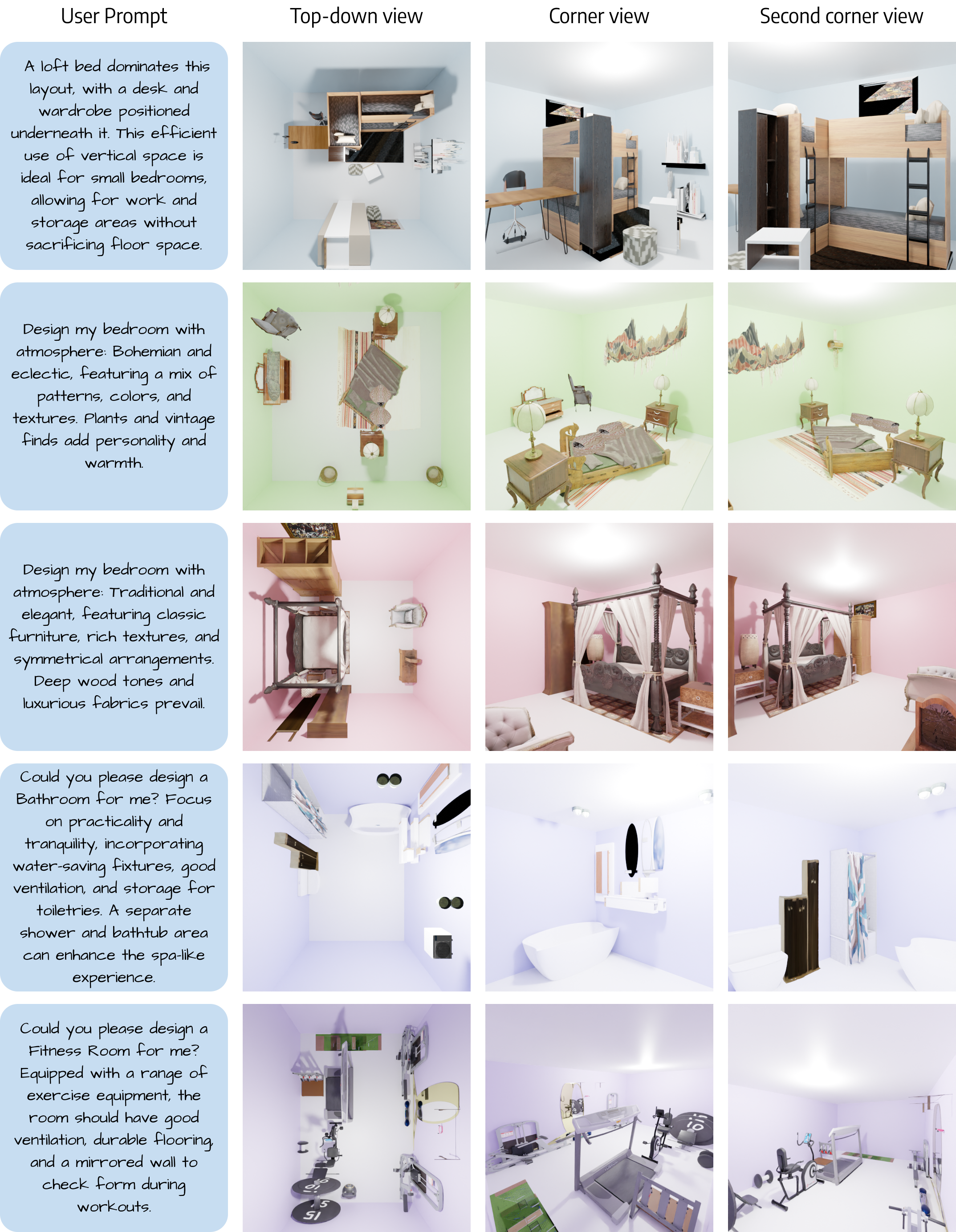}
  \caption{
    \textbf{Generating Renders through Elaborate Prompting}
    This collection presents rendered samples used for the evaluation in Tab.~\ref{tab:prompt_eval_unary} of the main paper. 
    The prompts for generating indoor scenes vary across five categories: Atmosphere, Scheme, Layout, and Functionality.
  }
  \label{fig:supp:gallery_w_prompt}
\end{figure}

\begin{figure}[H]
  \centering
  \includegraphics[width=\textwidth,height=\textheight, keepaspectratio]{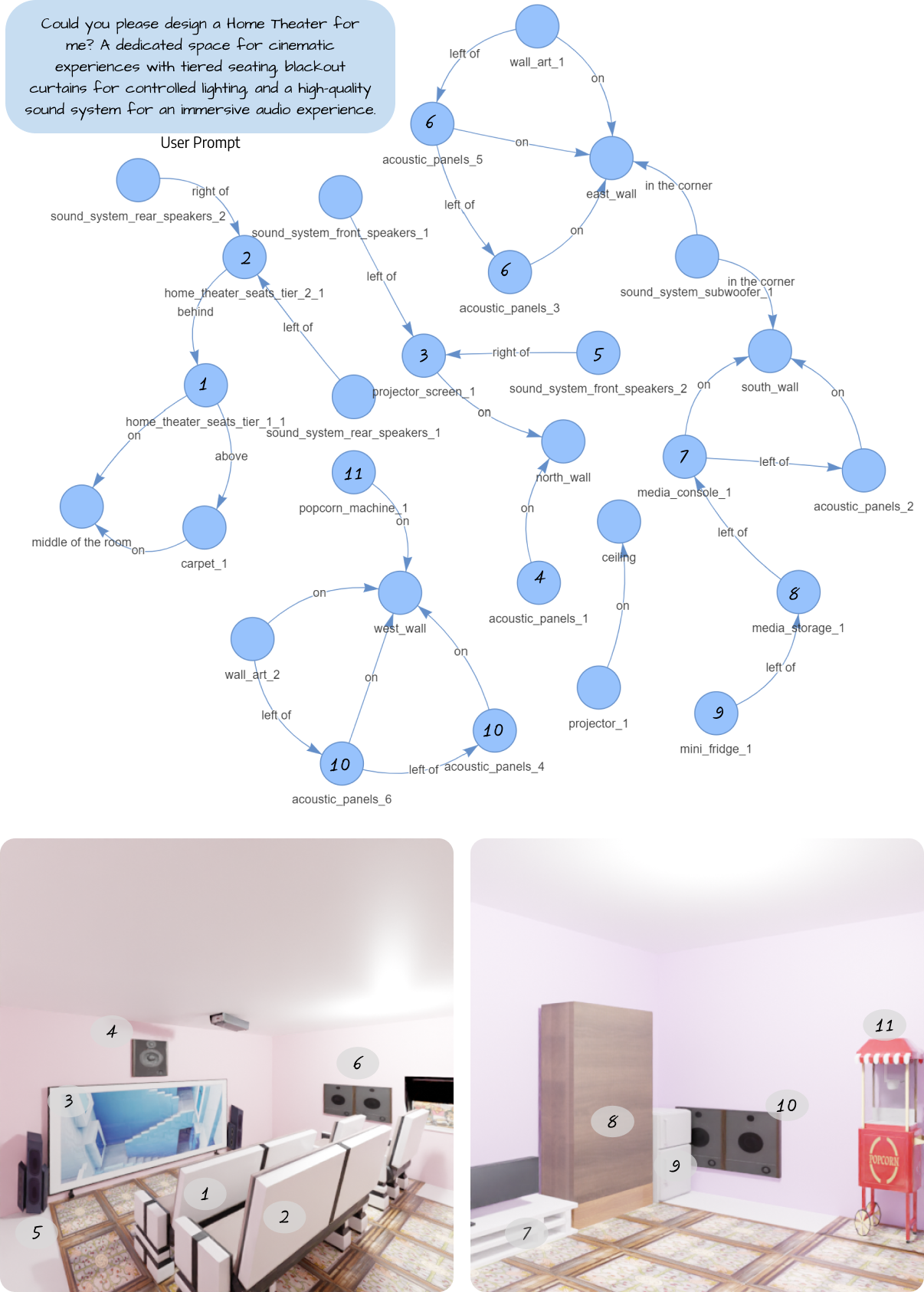}
  \caption{
    \textbf{Synthesized Scene Graph and Corresponding Renders}
    We illustrate the transformation of a user prompt into a scene graph and then into its corresponding 3D representation.
  }
  \label{fig:supp:scenegraph}
\end{figure}

In this section, we present additional qualitative outcomes. Fig.~\ref{fig:supp:gallery} shows a range of living room renders generated using generic prompts. 
Fig.~\ref{fig:supp:gallery_w_prompt} highlights our method's capability in handling various prompt types. 
Fig.~\ref{fig:supp:scenegraph} illustrates the conversion of a generated scene graph into a 3D representation.

\section{Implementation of the Multi-agent Pipeline}
As stated in Sec.~\ref{sec:llm-agent}, the responsibility of the Multi-agent pipeline includes suggesting scene objects, determining their relative placements, and structuring this information into a JSON object that conforms to a specified schema. There are a few motivations for distributing these tasks to multiple LLM instances. 

The main reason for following a ``Divide \& Conquer'' approach is the observed performance increase when each LLM instance focuses on solving a more trivial sub-problem by attending to the outputs of the previous agents during communication in an attempt to solve the entirety of the problem iteratively. Adding a feedback loop to the communication process further helps erase unwanted behavior from the agents and tackle hallucination issues.

Additionally, task distribution between agents addresses another challenge: not surpassing the output token limit of the LLM when dealing with a large number of objects. Currently, GPT-4 supports an output token limit of 4000 tokens, which constrains the number of objects and information that an LLM instance can handle in a single iteration. Furthermore, we observed that as LLM outputs approach the token limit, the attention to user and system prompts drastically decreases, resulting in object suggestions or placements that do not align with user preferences. Distributing the information processing among multiple agents mitigates this problem and allows more objects to be successfully generated.

In our implementation, the Interior Designer and Interior Architect agents generate and process all objects together in one pass. This design choice enables the agents to attend to each object effectively. We additionally explored alternative approaches wherein objects would be suggested and placed incrementally as the scene is completed. However, our experiments revealed repetitive object suggestion patterns with these alternative schemes, with some objects repeatedly proposed multiple times. This suggests that as the number of objects in the scene increases, the agents may struggle to comprehend the semi-completed scenes effectively. All GPT-4 agents utilize the ``JSON mode,'' which exclusively restricts their output to the JSON format. This approach further reduces the number of tokens spent in each iteration.

As the Engineer's task solely involves fitting the outputs of the Interior Designer and Architect agents into a specified schema, the Engineer agent can, in contrast, generate structured outputs for each object individually. This step is independent of each object and does not require consideration of other objects in the scene. Due to the aforementioned limitations regarding large output chunks, we opt for an iterative structuring process for the Engineer agent.

\section{Implementation of the GPT-4V Evaluator}
In Sec.~\ref{sec:quantitative}, we presented a grading-oriented evaluation approach leveraging the GPT-4V model. This system enables the measurement and comparison of abstract concepts like atmosphere and color schemes using renders from the scenes. As reflected in Sec.~\ref{sec:user_study}, the findings from the user study align with the GPT-4V evaluation, demonstrating the effectiveness of our LLM-based evaluation method. 

In our current experimental setup, we encode two separate renders of a scene from opposing corners and prompt the GPT-4V evaluator to grade the scene in four categories: Atmosphere, Scheme, Layout, and Functionality. The evaluator is executed three times in total, and the mean grade is calculated to reduce the variability in grades across different runs. 

To stimulate a ``chain of thought'' process~\cite{wei2022chain}, we instructed the GPT-4V evaluator to comment on the scenes regarding various grading aspects before generating the actual grades. This approach enables the model to consider its comments throughout the grading process.

\subsection{GPT-4V Evaluator System Prompt}

Give a grade from 0 to 10 to the following room renders based on how well they correspond together to the user preference (in triple backquotes) in the following aspects:
\\\\
- Realism and 3D Geometric Consistency
\\
- Functionality and Activity-based Alignment
\\
- Layout and furniture  
\\
- Color Scheme and Material Choices
\\
- Overall Aesthetic and Atmosphere   
\\\\
User Preference:
``\textbf{\{prompt\}}''
\\\\
Return the results in the following JSON format:
``\textbf{\{example\_json\}}''

\section{Discussion}

In this section, we delve into the existing limitations of our pipeline and potential future research directions.

One limitation stems from the LLM's inability to consider multiple factors simultaneously (including the rotation of the parent object, placement of other children, availability of space, and so forth) when integrating an object into a scene, particularly as the scene becomes more complex.
Thus, as the scene graph grows, the likelihood of the LLM suggesting unreasonable relational edges among objects increases. Likewise, as the number of objects in the scene increases, LLMs struggle to monitor the remaining available space, which may result in an overflow of objects in certain parts of the scene.

Another issue arises due to the discrepancy between the bounding box scale ratios inferred by the LLM and the actual asset retrieved. A laptop is a good example of such a case since closed and open laptops have very different bounding box ratios. If the agent infers a closed laptop, but the retrieved asset is an open laptop, the renders may result in visually implausible outcomes with distortions because of the resizing process. A similar discrepancy might also arise for object orientations. The canonical orientation of an asset might differ from the canonical orientation inferred by agents. For example, determining the front side of a desk can be subjective. The front side might be considered where a person would typically place a chair, or it could be perceived as the opposite side.

To maintain a consistent and immersive atmosphere, cohesive texture becomes essential. However, the original textures of retrieved objects from the dataset cannot be guaranteed to align seamlessly with user preference. Achieving precise control over texture and geometry-related features remains challenging despite extracting assets from large-scale datasets guided by text embeddings to ensure semantic alignment. Future work may consider incorporating a re-texturing step to enhance the coherence of the overall atmosphere. 

Lastly, we place objects in the scene with a bounding box assumption, meaning that the objects are assumed to have quadratic surfaces. While this approach simplifies the spatial representation of 3D objects and prevents collisions, it can lead to spatial inconsistencies when placing child objects, especially when placing a child object on top of a parent object. This limitation may result in artifacts such as ``floating'' objects. For further work, we encourage exploring an extra step that utilizes mesh surfaces instead of bounding boxes for precise placement.      

\section{System Prompts for Agents}

Creating the scene graph involves various agents, each playing a distinct role within the overall generation process. These roles are imparted to the agents through ``system prompts,'' shaping their responses to user prompts and instilling them with individual characteristics. These prompts help define each agent's functionality and guide them in generating structured outputs. The system prompts for each agent are provided verbatim below. The JSON schemas provided to each agent (highlighted with \textbf{\{json\_schema\}}) are included in the supplementary files and the project webpage.

\subsection{Interior Agent System Prompt}

Interior Designer. Suggest \{n\} essential new objects to be added to the room based on the user preference, the general functionality of the room, and the room size. The suggested objects should contain the following information:
\\\\
1. Object name (ex., bed, desk, chair, monitor, bookshelf, etc.)
\\\\
2. Architecture style (ex., modern, classic, etc.)
\\\\
3. Material (ex., wood, metal, etc.)
\\\\
4. Bounding box size in meters (ex., Length: 1.0m, Width: 1.0m, Height: 1.0m). Only use ``Length'', ``Width'', and ``Height'' as keys for the bounding box size.
\\\\
5. Quantity (ex., 1, 2, 3, etc.)
IMPORTANT: Do not suggest any objects related to doors or windows, such as curtains, blinds, etc.
\\\\
Follow the JSON schema below:
\textbf{\{json\_schema\}}

\subsection{Interior Architect System Prompt}
Interior Architect. Your role is to analyze user preferences, consider the optimal placement for each object that the Interior Designer suggests, find a place for this object in the room, and give a detailed description of it.
If the quantity of an object is greater than one, you have to find a place for each instance of this object separately but give all this information in one list item.
Give explicit answers for EACH object on the following three aspects:
\\\\
Placement: 
Find a relative place for the object (e.g., in the middle of the floor, in the northwest corner, on the east wall, right of the desk, on the bookshelf).
For relative placement with other objects in the room, use the prepositions ``on'', ``left of'', ``right of'', ``in front'', ``behind'', ``under''.
For relative placement with the room layout elements (walls, the middle of the room, ceiling), use the prepositions ``on'' and ``in the corner''.
You are not allowed to use any prepositions other than the ones above. Explicitly state the placement for each instance (ex., one is on the left of desk\_1, one is on the south\_wall).
\\\\
Proximity : 
Proximity of this object to the relative placement objects:
\\
1. Adjacent: The object is physically contacting the other object, or the other object supports it, or they are touching, or they are close to each other.
\\
2. Not Adjacent: The object is not physically contacting the other object and is distant from it.
\\\\
Facing :
Think about which wall (west/east/north/south\_wall) this object should be facing and explicitly state this (ex., one is facing the south\_wall, one is facing the west\_wall).
\\\\
If the quantity of an object is greater than one, you have to find a place for each instance of this object separately but give all this information in one list item.
\\\\
Follow the JSON schema below:
\textbf{\{json\_schema\}}

\subsection{Engineer System Prompt}
Engineer. You listen to the input by the Admin and create a JSON file.
\\\\
When the Admin outputs objects to be in the room, you will save ALL of them in the given schema.
For the scene graph, you can use the IDs for the objects already in the room but only output the objects to be placed.
If an object has a quantity higher than one, save each instance of this object separately.
\\\\
IMPORTANT: The inputted ``Placement'' key should be used for the ``placement'' key in the JSON object. Follow exactly the prepositions stated; do not use the information in the ``Facing'' key for the room layout elements.
\\\\
IMPORTANT: For object quantities greater than one, the ``placement'' key gives separately the relative placement of each instance of that object in the room; make the distinction for each instance accordingly.
\\\\
Use only the following JSON Schema to save the JSON object:
\textbf{\{json\_schema\}}

\subsection{Layout Corrector System Prompt}
Layout Corrector Agent. Whenever a user provides an object that doesn't fit the room for various spatial conflicts, you will change its ``scene\_graph'' and ``facing\_object'' keys to resolve these conflicts. 
\\\\
You will use the JSON Schema to validate the user's JSON object.
\\\\
For relative placement with other objects in the room, use the prepositions ``on'', ``left of'', ``right of'', ``in front'', ``behind'', ``under''.
For relative placement with the room layout elements (walls, the middle of the room, ceiling), use the prepositions ``on'', and ``in the corner''.
\\\\
Use only the following JSON Schema to save the JSON object:
\textbf{\{json\_schema\}}

\subsection{Layout Refiner System Prompt}
Layout Refiner. Whenever the Admin speaks, you will look at the parent object and children objects, the first preposition that connects these objects, and find a second suitable relative placement for the children objects while considering the initial positioning of the object. 
Give the relative placement of the children objects with each other and with the parent object. For example, if there are five children objects ``on'' the parent object, give the relative positions of the children objects to one another and the second preposition to the parent object (``on'' is the first preposition).
\\\\
Use only the following JSON Schema to save the JSON object:
\textbf{\{json\_schema\}}

\section{Room Synthesis Prompt Generation for Evaluation}
Our proposed pipeline aims to serve as a personalized interior design assistant capable of offering tailored design suggestions based on diverse user inputs, reflecting various preferences and requirements. Our motivation is to construct prompts that reflect the authentic comments real people might make when designing their rooms. 

In the realm of interior design, the significance of both aesthetics and functionality is consistently emphasized~\cite{ching2018interior}. A deeper exploration of personalized interior design, as articulated in~\cite{lee2022conceptual}, identifies four pivotal aspects central to decision-making: indoor space components, human tendencies, technology, and spatial evaluation. Indoor space components encompass factors such as materials, layout, and arrangement, while human tendencies consider individual inclinations influenced by past experiences and personalities. Spatial evaluation accentuates productivity, underlining the importance of creating environments conducive to efficient work.

Drawing from this framework, we establish our GPT-4V evaluation criterion and generate prompts by varying key elements, including functionality, layout, color scheme with material choice, and overall atmosphere. These prompts are generated utilizing GPT-4. For instance, in generating prompts related to functionality, we instruct GPT-4 to provide descriptions for different room types, incorporating varied functionalities and dimension specifications. Similarly, for prompts assessing layout, color scheme, and overall atmosphere alignment, we maintain fixed room types and dimensions while altering layout, color scheme, or atmosphere. We generate ten prompts for each alignment consideration. Below are the prompts for generating room preferences and a comprehensive list of prompts used in the evaluations~\cref{tab:quant_eval} and \cref{tab:prompt_eval_unary} of the main paper.

\subsection{Prompts for Room Preferences Text Generation} 
In the evaluation recorded in \cref{tab:quant_eval}, we utilize minimal preference descriptions as prompts for room synthesis. The prompt for generating these preferences is noted in the section titled ``Minimal Preference Generation Prompt'' below. The preferences generated for room synthesis evaluation in \cref{tab:prompt_eval_unary} of the main paper are listed in the other sections.
\\\\
\noindent
\textit{\textbf{Minimal Preference Generation Prompt}} \\
\textit{Bedroom: }\\
You are a helpful assistant.\\
Could you please provide a list of common dimensions for a bedroom? Please list ten potential dimensions, including width, length, and height in meters. Format your response as follows: [[width, length, height], [width, length, height], ...]
\\\\
\textit{Livingroom: }\\
You are a helpful assistant. \\
Could you please provide a list of common dimensions for a living room? Please list ten potential dimensions, including width, length, and height in meters. Format your response as follows: [[width, length, height], [width, length, height], ...]
\\\\
\noindent
\textit{\textbf{Functionality-related Preference Generation Prompt}}\\
You are a helpful assistant who is designed to output JSON. \\
Please provide ten interior design instructions emphasizing functionality. \\
Begin by specifying the room type and desired room dimensions in meters, including width, length, and height. Describe the room's intended functionality succinctly. Besides, common suggestions also extend to more diverse requirements, such as creating a reading corner or a movie area. \\
Keep descriptions brief, within three sentences, and exclude details involving windows and doors. Aim for diversity in interior room types.\\
Provide the results in JSON format: \{``room type'': \{``dimension'': [width, length, height], ``functionality'': ``functionality description''\}, ``room type'': \{``dimension'': [width, length, height], ``functionality'': ``functionality description''\}, ...\} Note: you need to replace the key ``room type'' with specified room type, such as bedroom, living room etc. 
\\\\
\noindent
\textit{\textbf{Layout-related Preference Generation Prompt}}\\
You are a helpful assistant designed to output JSON.\\
Could you outline ten descriptions of potential bedroom interior design layouts? Please do not include descriptions about style, theme, etc, and only focus on layout.\\
Please present the results in JSON format as follows: \{``1'': ``layout and furniture description'', ``2'': ``layout and furniture description'', ...\}. Please keep each description concise with two to three sentences.
\\\\
\noindent
\textit{\textbf{Color-scheme-and-material-related Preference Generation Prompt}}\\
You are a helpful assistant designed to output JSON.\\
Can you list ten interior bedroom design requirements or ideas with different color themes (e.g., room with pink color colors or beige tones) and material emphasis (e.g., room with wooden elements or metallic elements)? Please do not use windows, ceilings, floors, and walls in the descriptions. \\
Provide the results in JSON format: \{``1'': ``requirement description'', ``2'': ``requirement description'', ...\}. Please keep the description short, with no more than four sentences.
\\\\
\noindent
\textit{\textbf{Overall-atmosphere-related Preference Generation Prompt}}\\
You are a helpful assistant designed to output JSON.\\
Can you provide ten concise descriptions of the overall atmosphere for potential bedroom interior designs? \\
Please present the results in JSON format as follows: \{``1'': ``atmosphere description'', ``2'': ``atmosphere description'', ...\}. Please keep each description concise within two to three sentences.

\subsection{Complete Lists of Prompts}
The comprehensive prompts list for room synthesis evaluations documented in \cref{tab:quant_eval} of the main paper is provided in Tab.~\ref{tab:promtlist_minimal} below. Similarly, the prompts for evaluations recorded in \cref{tab:prompt_eval_unary} are detailed in Tab.~\ref{tab:promtlist_others}.

\begin{longtable}{
    >{\arraybackslash}p{2cm} 
    >{\arraybackslash}p{6cm} 
    >{\centering\arraybackslash}p{4cm}
}
\caption{\textbf{Complete List of Prompts for Evaluation Recorded in \cref{tab:quant_eval} of the main paper.} Prompts 1 to 10 describe \textit{bedrooms};  11 to 20 describe \textit{living rooms}.} \\
\label{tab:promtlist_minimal} \\
\toprule
\textbf{Index} & \textbf{Prompt} & \textbf{Room Dimension} \\
\midrule
\endfirsthead
\multicolumn{3}{c}{{\bfseries \tablename\ \thetable{} -- Continued from previous page}} \\
\toprule
\textbf{Index} & \textbf{Prompt} & \textbf{Room Dimension} \\
\midrule
\endhead
\bottomrule
\multicolumn{3}{c}{{\bfseries Continued on next page}}
\endfoot
\bottomrule
\endlastfoot
1 & Design me a bedroom. & [3.0, 4.0, 2.4] \\
2 & Design me a bedroom. & [2.5, 3.0, 2.4] \\
3 & Design me a bedroom. & [3.5, 4.5, 2.4] \\
4 & Design me a bedroom. & [4.0, 5.0, 2.4] \\
5 & Design me a bedroom. & [2.4, 3.5, 2.4] \\
6 & Design me a bedroom. & [3.2, 4.2, 2.4] \\
7 & Design me a bedroom. & [2.8, 3.6, 2.4] \\
8 & Design me a bedroom. & [3.6, 4.8, 2.4] \\
9 & Design me a bedroom. & [4.2, 5.2, 2.4] \\
10 & Design me a bedroom. & [3.0, 3.5, 2.4] \\
11 & Design me a living room. & [4.0, 5.0, 2.8] \\
12 & Design me a living room. & [3.5, 4.5, 2.8] \\
13 & Design me a living room. & [3.0, 4.0, 2.8] \\
14 & Design me a living room. & [4.5, 6.0, 3.0] \\
15 & Design me a living room. & [5.0, 7.0, 3.0] \\
16 & Design me a living room. & [3.6, 4.8, 2.8] \\
17 & Design me a living room. & [4.2, 5.2, 2.8] \\
18 & Design me a living room. & [5.5, 6.5, 3.0] \\
19 & Design me a living room. & [3.2, 4.2, 2.8] \\
20 & Design me a living room. & [6.0, 8.0, 3.0] 

\end{longtable}

\begin{longtable}{
    >{\arraybackslash}p{1.2cm} 
    >{\arraybackslash}p{8cm} 
    >{\centering\arraybackslash}p{2.5cm}
}
\caption{\textbf{Complete List of Prompts for Evaluation Recorded in \cref{tab:prompt_eval_unary} of the main paper.} Prompts from index 1 to 10 pertain to \textit{functionality} preferences, those from index 11 to 20 concern \textit{layout}, index 21 to 30 relate to \textit{color scheme and materials}, and prompts from index 31 to 40 encompass \textit{overall atmosphere} preferences.} \\
\label{tab:promtlist_others} \\
\toprule
\textbf{Index} & \textbf{Prompt} & \textbf{Room Dimension} \\
\midrule
\endfirsthead
\multicolumn{3}{c}{{\bfseries \tablename\ \thetable{} -- Continued from previous page}} \\
\toprule
\textbf{Index} & \textbf{Prompt} & \textbf{Room Dimension} \\
\midrule
\endhead
\bottomrule
\multicolumn{3}{c}{{\bfseries Continued on next page}}
\endfoot
\bottomrule
\endlastfoot
1 & Could you please design a Living Room for me? Designed for relaxation and socializing, the living room should feature comfortable seating areas, ample lighting for different activities, and a designated movie area with a large screen and surround sound for an immersive experience. & [6, 8, 2.5] \\
2 & Could you please design a Home Office for me? The home office should prioritize a clutter-free workspace with ergonomic furniture, ample storage for office supplies, and a small area for breaks with a comfortable chair and a coffee machine. & [3, 4, 2.5] \\
3 & Could you please design a Kitchen for me? Efficiency and ease of movement are key, with a triangular layout between the stove, refrigerator, and sink. Include a central island for additional workspace and seating for casual dining. & [5, 7, 2.5] \\
4 & Could you please design a Bedroom for me? A cozy and restful environment with a comfortable bed, soft lighting, and ample storage for personal items. A small reading nook with a comfy chair and a bookshelf should be included. & [4, 5, 2.5] \\
5 & Could you please design a Bathroom for me? Focus on practicality and tranquility, incorporating water-saving fixtures, good ventilation, and storage for toiletries. A separate shower and bathtub area can enhance the spa-like experience. & [3, 4, 2.5] \\
6 & Could you please design a Dining Room for me? Designed for meal gatherings, it should have a large table with comfortable seating for the family and guests, along with ambient lighting to enhance the dining experience. & [4, 6, 2.5] \\
7 & Could you please design a Playroom for me? A vibrant and flexible space that encourages play, creativity, and learning, with durable, easy-to-clean surfaces, storage for toys, and a comfortable area for reading and crafts. & [4, 6, 2.5] \\
8 & Could you please design a Fitness Room for me? Equipped with a range of exercise equipment, the room should have good ventilation, durable flooring, and a mirrored wall to check form during workouts. & [4, 6, 2.5] \\
9 & Could you please design a Laundry Room for me? A functional space with efficient appliances, a fold-out ironing board, and storage for cleaning supplies to make laundry tasks easier and organized. & [3, 3, 2.5] \\
10 & Could you please design a Home Theater for me? A dedicated space for cinematic experiences with tiered seating, blackout curtains for controlled lighting, and a high-quality sound system for an immersive audio experience. & [5, 7, 2.5] \\
11 & Design my bedroom with following layout: This layout features a queen-sized bed against the main wall, with two nightstands on either side. Opposite the bed, there's a dresser with a mirror above it, creating a functional dressing area. & [4, 4, 2.5] \\
12 & Design my bedroom with following layout: In this setup, a single bed is placed in the corner, maximized for space efficiency. A small desk and chair fit snugly in the opposite corner, with a tall bookshelf beside it, making it ideal for a student. & [4, 4, 2.5] \\       
13 & Design my bedroom with following layout: This layout utilizes a king-sized bed centered on the main wall, with a bench at its foot. A large wardrobe is placed on the adjacent wall, providing ample storage space without cluttering the room. & [4, 4, 2.5] \\
14 & Design my bedroom with following layout: A twin bed is positioned against one wall, leaving space for a play area on the opposite side of the room. Toy storage and a small table with chairs are included in the play area, perfect for children. & [4, 4, 2.5] \\
15 & Design my bedroom with following layout: The room features a full bed flanked by a desk on one side and a nightstand on the other. Across from the bed, a low media console serves as a place for entertainment equipment, optimizing the layout for relaxation and study. & [4, 4, 2.5] \\
16 & Design my bedroom with following layout: In this compact layout, a murphy bed is installed to maximize floor space when not in use. A fold-down desk is mounted on the opposite wall, creating a multipurpose space that can easily transition from bedroom to home office. & [4, 4, 2.5] \\
17 & Design my bedroom with following layout: A loft bed dominates this layout, with a desk and wardrobe positioned underneath it. This efficient use of vertical space is ideal for small bedrooms, allowing for work and storage areas without sacrificing floor space. & [4, 4, 2.5] \\
18 & Design my bedroom with following layout: This spacious layout includes a queen-sized bed positioned centrally with a vanity table and stool set against the adjacent wall. A comfortable reading chair and floor lamp are placed in one corner, creating a cozy reading nook. & [4, 4, 2.5] \\
19 & Design my bedroom with following layout: Featuring a platform bed with storage drawers beneath, this layout optimizes storage. Along the opposite wall, a long, low dresser doubles as a display surface for personal items, with a mirror above it to enhance natural light. & [4, 4, 2.5] \\
20 & Design my bedroom with following layout: In this innovative layout, the bed is centrally located with a headboard that doubles as a room divider. Behind the headboard, a workspace is created with a desk and shelving, effectively separating the sleeping and working areas. & [4, 4, 2.5] \\
21 & Design a bedroom with color theme Minimalist with White Tones for me. Use white and light grey hues to emphasize cleanliness and simplicity. Incorporate sleek, modern furniture with straight lines. Material focus on matte finishes and textiles like cotton or linen for a soft touch. & [4, 4, 2.5] \\
22 & Design a bedroom with color theme Boho Chic with Earthy Tones for me. Focus on mixing patterns, colors, and textures. Use materials like rattan, bamboo, and unfinished woods. Incorporate plants and macrame textiles for a cozy, natural feel. & [4, 4, 2.5] \\
23 & Design a bedroom with color theme Industrial with Metallic Elements for me. Incorporate exposed steel, iron, or brushed nickel finishes in decor items and furniture. Use a neutral palette with bold accents in art or textiles. Emphasize raw and unfinished looks. & [4, 4, 2.5] \\
24 & Design a bedroom with color theme Modern Glam with Gold Accents for me. Use a base of neutral colors with pops of bold color. Integrate gold-trimmed furniture and gold accent decor for a touch of luxury. Velvet and silk fabrics add texture and opulence. & [4, 4, 2.5] \\ 
25 & Design a bedroom with color theme Nautical with Blue Colors for me. Incorporate various shades of blue with crisp white for a sea-inspired look. Use striped patterns and nautical decor items. Materials include weathered wood and rope accents for a maritime feel. & [4, 4, 2.5] \\
26 & Design a bedroom with color theme Scandinavian with Pastel Colors for me. Use pale blues, pinks, and greens on a backdrop of white or grey. Furniture is minimalist and functional, with natural light wood materials. Add cozy textiles like wool or mohair to enhance comfort. & [4, 4, 2.5] \\
27 & Design a bedroom with color theme Rustic with Wooden Elements for me. Emphasize natural, unfinished woods in furniture and decor for a warm, earthy feel. Use leathers and woven textiles to add depth. Incorporate organic, handmade items to underscore the rustic theme. & [4, 4, 2.5] \\
28 & Design a bedroom with color theme Art Deco with Rich Jewel Tones for me. Combine deep greens, blues, and purples with metallic accents in gold or brass. Use geometric patterns in textiles and art. Furniture and decor pieces should evoke the luxury and opulence of the 1920s and 1930s. & [4, 4, 2.5] \\
29 & Design a bedroom with color theme Contemporary with Monochromatic Scheme for me. Stick to a monochromatic color scheme throughout, using varying shades of the same color for depth. Focus on sleek furniture with minimalist designs. Utilize textures like glass, polished metals, and smooth fabrics to add interest. & [4, 4, 2.5] \\
30 & Design a bedroom with color theme Vintage with Floral Patterns for me. Incorporate floral patterns in textiles, art, and wallpaper. Use a mix of antique or vintage-style furniture with rich wood tones. Embrace lace and embroidered textiles for a delicate, classic touch. & [4, 4, 2.5] \\
31 & Design my bedroom with atmosphere: Minimalist and serene, with clean lines and a monochrome palette. Accentuated by natural light and a lack of clutter. & [4, 4, 2.5] \\
32 & Design my bedroom with atmosphere: Bohemian and eclectic, featuring a mix of patterns, colors, and textures. Plants and vintage finds add personality and warmth. & [4, 4, 2.5] \\
33 & Design my bedroom with atmosphere: Modern and sleek, characterized by bold geometric shapes and a neutral color scheme. Innovative lighting and high-tech elements are key. & [4, 4, 2.5] \\
34 & Design my bedroom with atmosphere: Cozy and rustic, emphasizing natural wood, stone, and warm, earthy tones. Chunky knits and a fireplace complete the inviting ambiance. & [4, 4, 2.5] \\
35 & Design my bedroom with atmosphere: Nautical and breezy, with a color palette of blues, whites, and sandy tones. Maritime accessories and striped patterns evoke the seaside. & [4, 4, 2.5] \\
36 & Design my bedroom with atmosphere: Glamorous and luxurious, marked by opulent fabrics, metallic finishes, and a touch of sparkle. Elegant furniture and plush textiles dominate. & [4, 4, 2.5] \\
37 & Design my bedroom with atmosphere: Industrial and edgy, with exposed brick, metal details, and raw concrete elements. A neutral color scheme is offset by vibrant art. & [4, 4, 2.5] \\
38 & Design my bedroom with atmosphere: Traditional and elegant, featuring classic furniture, rich textures, and symmetrical arrangements. Deep wood tones and luxurious fabrics prevail. & [4, 4, 2.5] \\
39 & Design my bedroom with atmosphere: Scandinavian and bright, with a focus on simplicity, functionality, and minimalism. Pale woods, muted colors, and hygge accents are key. & [4, 4, 2.5] \\
40 & Design my bedroom with atmosphere: Contemporary and dynamic, with a mix of textures and materials. Clean lines, pops of color, and versatile pieces adapt to changing trends. & [4, 4, 2.5] 

\end{longtable}

\end{document}